%% file: main.tex
\documentclass[10pt,twocolumn,letterpaper]{article}

\usepackage[pagenumbers]{iccv} 

\usepackage{makecell}

\usepackage{amsmath}
\usepackage{bm}
\usepackage{amsmath,amsfonts}
\usepackage{algorithmic}
\usepackage{algorithm}
\usepackage{gensymb}

\usepackage{tcolorbox}

\definecolor{iccvblue}{rgb}{0.21,0.49,0.74}
\usepackage[pagebackref,breaklinks,colorlinks,allcolors=iccvblue]{hyperref}

\def\paperID{11409} 
\def\confName{ICCV}
\def\confYear{2025}

\title{MuMA: 3D PBR Texturing via Multi-Channel Multi-View Generation \\ and Agentic Post-Processing}
\author{
Lingting Zhu$^{1}$\thanks{Equal contribution.} \quad Jingrui Ye$^{2}$\footnotemark[1] \quad Runze Zhang$^{3}$ \quad Zeyu Hu$^{3}$ \quad Yingda Yin$^{3}$ 
\quad Lanjiong Li$^{4}$ \\ \quad Jinnan Chen$^{5}$ \quad Shengju Qian$^{3}$ \quad Xin Wang$^{3}$ \quad Qingmin Liao$^{2}$\thanks{Corresponding author.} \quad  Lequan Yu$^{1}$\footnotemark[2]\\
$^1$HKU $^2$Tsinghua SIGS $^3$LIGHTSPEED $^4$HKUST(GZ) $^5$NUS
}

\begin{document}
\maketitle

\def\@fnsymbol#1{\ensuremath{\ifcase#1\or *\or \dagger\or  \ddagger\or
  \mathsection\or \mathparagraph\or \|\or **\or \dagger\dagger
  \or \ddagger\ddagger \else\@ctrerr\fi}}
\renewcommand{\thefootnote}{\fnsymbol{footnote}}

\input{sec/abstract}
\input{sec/1_intro}
\input{sec/2_related}
\input{sec/3_method}

\input{sec/4_exp}

\section{Conclusion}
In this work, we present MuMA, a novel framework designed to produce high-quality 3D texturing with physically based rendering (PBR). Following the recent fashion of first generating multi-view images and then performing UV unwrapping, we propose to model the generation of shaded and albedo images. This process produces an albedo map and connects to intrinsic decomposition models for the remaining PBR channels. The key insight behind this approach is to simplify multi-channel training and mitigate the issue of unbalanced data. Additionally, our framework integrates post-processing that leverages multimodal language models to score and select the best results, demonstrating an agentic system that mimics the behavior of 3D artists. Our method outperforms existing research in 3D PBR texturing, and we anticipate that MuMA contributes to the evolving landscape of 3D content creation.
\input{sec/X_suppl}

\clearpage
{
    \small
    \bibliographystyle{ieeenat_fullname}
    \bibliography{main}
}


\end{document}

%% file: sec/abstract.tex
\begin{abstract}
Current methods for 3D generation still fall short in physically based rendering (PBR) texturing, primarily due to limited data and challenges in modeling multi-channel materials. In this work, we propose \textbf{MuMA}, a method for 3D PBR texturing through Multi-channel Multi-view generation and Agentic post-processing. Our approach features two key innovations: 1) We opt to model shaded and albedo appearance channels, where the shaded channels enables the integration intrinsic decomposition modules for material properties. 2) Leveraging multimodal large language models, we emulate artists' techniques for material assessment and selection. Experiments demonstrate that MuMA achieves superior results in visual quality and material fidelity compared to existing methods.

\end{abstract}

%% file: sec/1_intro.tex
\section{Introduction}
\label{sec:intro}
3D imagination, bolstered by advancements in 3D generative modeling, significantly enhances human creativity and real-world design processes. In the realm of 3D generation, texturing is crucial for two main reasons. First, painting 3D models is a standard practice in design pipelines, where artists initially produce untextured models for downstream applications. Second, recent progress in 3D generation~\cite{zhang2024clay, zhao2025hunyuan3d, bensadoun2024meta} indicates that the decomposition of geometry and texture generation is becoming increasingly promising. However, despite the growing success of numerous 3D object generation methods~\cite{poole2022dreamfusion, liu2023zero, liu2023syncdreamer, long2024wonder3d, yang2024hunyuan3d, hong2023lrm}, progress in 3D texturing, particularly in modeling physically based rendering (PBR) materials, remains limited due to data constraints and unresolved material modeling challenges.

\begin{figure}[t]
\centering
\includegraphics[width=0.48\textwidth]{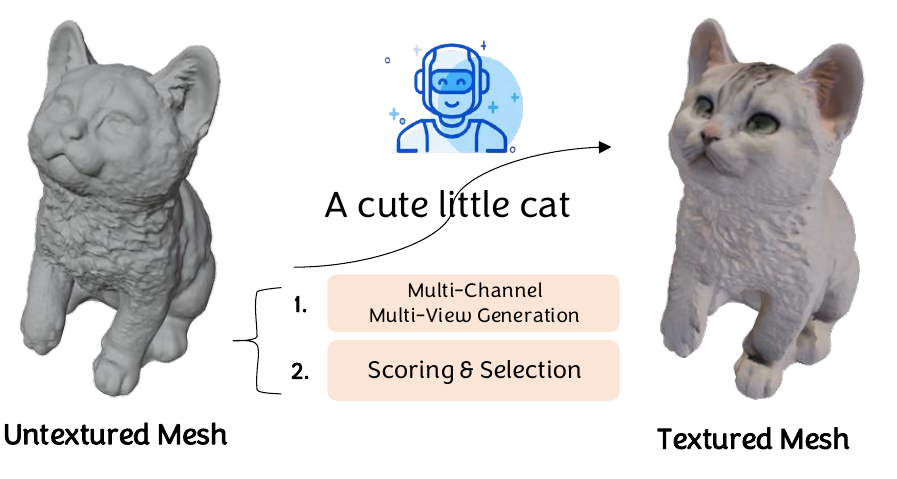}
\caption{\textbf{Illustration of our method.} Given an untextured mesh, our method produces PBR texuring from user inputs with multi-channel multi-view generation and agentic post-processing.} 
\label{teaser}
\end{figure}

Earlier works on 3D texturing leverage 2D priors from pre-trained diffusion models~\cite{ho2020denoising, rombach2022high}. These approaches can be categorized into two main streams. The first stream~\cite{chen2023fantasia3d, liu2025unidream, qiu2024richdreamer, zhang2024dreammat} adopts the differentiable optimization method of Score Distillation Sampling~\cite{poole2022dreamfusion}, generating textures by minimizing the noise prediction loss across various views. The second stream~\cite{richardson2023texture, chen2023text2tex, liu2024text, zhang2024texpainter, cao2023texfusion} involves painting each viewpoint using diffusion models with geometry control branches~\cite{zhang2023adding}, combined with a progressive inpainting strategy. However, these methods often suffer from poor multi-view consistency and low texture quality.
More recently, to better address multi-view consistency, researchers~\cite{zeng2024paint3d, cheng2024mvpaint, zhao2025hunyuan3d, zhu2024mcmat, zhang2024clay} have employed multi-view diffusion to generate multi-view images and back-project them into the UV space, filling gaps through UV inpainting, thereby improving generation quality. Despite these advancements, the synthesis quality, particularly for PBR texturing, still falls short of human expectations. This limitation hinders downstream applications that require rendering under various lighting conditions.
In this work, we propose a method for 3D PBR texturing through Multi-channel Multi-view generation and Agentic post-processing, termed \textbf{MuMA}, illustrated in Fig.~\ref{teaser}. Our work features two rationales behind them, discussed in the following paragraphs. 

\begin{figure}[t]
\centering
\includegraphics[width=0.48\textwidth]{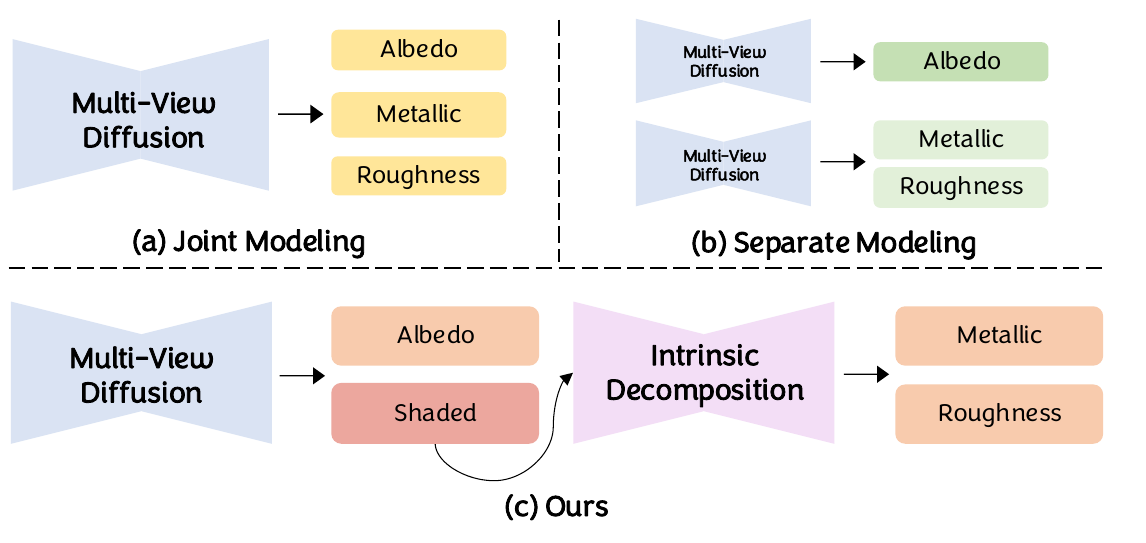}
\caption{\textbf{Illustration of different PBR modeling.} Unlike (a) Joint Modeling that faces training challenges and data imbalance, and (b) Separate Modeling encounters inconsistency between models, (c) Ours proposes to modeling albedo and shaded images to connect an intrinsic decomposition model.} 
\label{modeling_types}
\end{figure}

To ease multi-channel training and enable the acquisition of PBR parameters, we opt to model shaded and albedo appearance channels. The shaded channels allow for the integration of intrinsic decomposition modules for material properties. While existing works~\cite{zhang2024clay, zhu2024mcmat} have pioneered PBR generation using multi-channel diffusion modeling to jointly produce albedo, metallic, and roughness channels from text or image inputs, they fail to elegantly model multi-channel materials and produce high-fidelity PBR textures. This shortcoming is primarily due to two reasons. First, the domain gap between shaded/albedo images and metallic/roughness images hinders training, which is exacerbated when using base models like Stable Diffusion~\cite{rombach2022high}, trained on images under uncontrolled lighting conditions. Second, there is a significant imbalance in the availability of high-quality data for shaded/albedo images compared to metallic/roughness images, with the latter being scarce.
A straightforward approach might involve training two separate models for the different domains, but this can lead to inconsistencies between albedo and metallic/roughness channels. Recently, advancements in material decomposition methods~\cite{kocsis2024intrinsic, chen2024intrinsicanything, zeng2024rgb, li2024idarb, hong2024supermat} suggest that it is not necessary to achieve one-stage multi-channel PBR generation. Instead, a second-stage alternative can mitigate training dilemmas and dataset imbalances. Notably, IDArb~\cite{li2024idarb} addresses intrinsic decomposition for an arbitrary number of input views and illuminations, achieving multi-view consistent material decomposition and enabling a second-stage material decomposition.
Therefore, we model only shaded and albedo channels using a multi-view diffusion approach, built upon SDXL~\cite{podell2023sdxl} and MV-Adapter~\cite{huang2024mv}. The albedo images contribute to our final targets, while the shaded images connect our generative model to a pre-trained multi-view consistency material decomposition model, shown in Fig.~\ref{modeling_types}. It is worth noting that Meta AssetGen~\cite{siddiqui2025meta} also trains models to output shaded and albedo channels, but their pipeline and design serve different purposes. In our method, the multi-channel design is specifically used to connect to intrinsic decomposition modules and facilitate agentic post-processing.

Another key exploration of our work is the integration of agent systems. As 3D production in industrial settings requires significant expert-level human labor, it is crucial to mimic human behavior with autonomous agents to ensure high-quality creation. With the tremendous success of large language models (LLMs) and multimodal large language models (MLLMs)~\cite{brown2020language,gpt4o,liu2023visual}, agentic systems with strong reasoning and instruction-following capabilities pave the way for autonomous systems with proxy agents that can manipulate tools, access datasets, and refer to external information. There are works highlighting this demonstration in visual systems~\cite{shen2023hugginggpt,wu2023visual}, and earlier works in 3D generation~\cite{fang2024make,zhang2024mapa} attempt to apply MLLMs and LLMs for understanding and retrieval.
In our work, we apply an MLLM for scoring and selection, mimicking human behavior to produce the final version. Specifically, as the intrinsic decomposition model produces an alternative albedo map and we observe that sometimes these maps are competitive with our generated ones, we set up an auto-query mechanism to score and select between them, incurring nearly no external cost. Another common behavior in human production is the Best-of-N selection, which inspires an optional pipeline that draws multiple samples from our generated models and the pre-trained material decomposition model.

Our contributions are summarized as follows:
\begin{enumerate}
\item We propose a novel framework to generate high-fidelity physically based rendering textures. By designing a multi-channel, multi-view generation approach for shaded and albedo images, we enable high-quality training and the acquisition of PBR parameters.
\item We introduce an agentic framework to score and select candidates in both stages, facilitating multiple selections that mimic human behavior.
\item Extensive experiments demonstrate that our method excels in both appearance and material quality.
\end{enumerate}

\begin{figure*}[t] 
\centering
\includegraphics[width=0.95\textwidth]{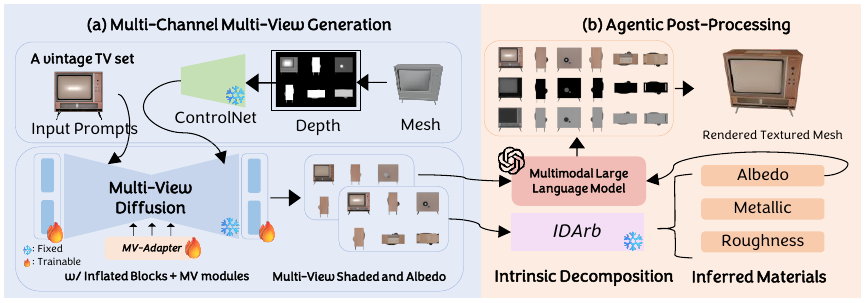} 
\caption{\textbf{Overview of MuMA.} \textbf{(a) Muti-Channel Multi-View Generation:} Given an untextured mesh and its description, we opt to model shaded and albedo images with Multi-View Diffusion. \textbf{(b) Agentic Post-Processing:} After producing the candidate material channels, we integrate multimodal languague models for scoring and selection, finally outputing the multi-view materials for texturing.}
\label{fig:framework}
\end{figure*}

%% file: sec/2_related.tex
\section{Related Works}
\label{sec:related}

\subsection{Texture Generation for 3D Object}
Given a 3D mesh without texture, TEXTure~\cite{richardson2023texture} proposes a progressive painting approach, leveraging a depth-conditioned diffusion model to generate textures one view at a time. Text2Tex~\cite{chen2023text2tex} introduces an automated view sequence generation strategy aimed at producing more consistent textures. Paint3D~\cite{zeng2024paint3d} back-projects generated front and back views onto a texture map and inpaints this map in the UV space, effectively filling gaps caused by self-occlusion. MVPaint~\cite{cheng2024mvpaint} uses four views to initialize the texture map and then outputs the final map through super-resolution technology and spatial-aware 3D inpainting. Hunyuan3D~\cite{zhao2025hunyuan3d} incorporates multi-view normal maps and position maps as control information, supporting the generation of images from more viewpoints and then back-projecting them to create a texture map.
Additionally, a series of works~\cite{zhang2024texpainter, liu2024text, huo2024texgen, gao2024genesistex, lu2024genesistex2} attempt to sample latent representations from multiple views concurrently to improve the global consistency of textures. Despite the significant progress achieved by these methods, the texture maps they generate are typically entangled with complex lighting and shadows, lacking realistic material modeling and hindering the integration into downstream rendering pipelines.

\subsection{Material Generation for 3D Object}
Earlier works have utilized Score Distillation Sampling (SDS) to optimize targets for PBR material generation. Fantasia3D~\cite{chen2023fantasia3d} introduces a spatially varying bidirectional reflectance distribution function (BRDF) into 3D generation to learn material properties. RichDreamer~\cite{qiu2024richdreamer} employs multi-view albedo data to train a diffusion model capable of generating multi-view albedos, but it does not account for metallic and roughness properties. Paint-it~\cite{youwang2024paint} uses convolution-based neural kernels to re-parameterize PBR texture maps, reducing noise impact during optimization. DreamMat~\cite{zhang2024dreammat} fine-tunes a diffusion model by pre-rendering 3D models under various environmental lighting, enabling it to predict PBR materials.
Some methods~\cite{zhang2024clay, huang2024material} train models to produce all PBR channels and use single-view or multi-view iterative back-projection, along with the RePaint~\cite{lugmayr2022repaint} algorithm, to obtain PBR maps. MaPa~\cite{zhang2024mapa} and other works~\cite{wang2024boosting, dang2024texpro} leverage large language models (LLMs) to guide material generation, improving the semantic accuracy of materials. TexGaussian~\cite{xiong2024texgaussian} proposes octant-aligned 3D Gaussian Splatting~\cite{kerbl20233d} for fast PBR material generation, though it may suffer from poor generalization.
However, all of these methods are either too time-consuming or prone to produce unnatural textures.

\subsection{Material Decomposition}
Material estimation methods aim to infer the intrinsic appearance of input images with unknown lighting and shadows. Some recent works~\cite{kocsis2024intrinsic, chen2024intrinsicanything, zeng2024rgb, li2024idarb, hong2024supermat} propose using diffusion models to capture the intrinsic material distribution of images. These models are designed to perceive lighting conditions and can efficiently predict intrinsic channels (albedo, metallic, roughness) on a pixel-by-pixel basis.
RGB$\ensuremath{\leftrightarrow}$X~\cite{zeng2024rgb} introduces a unified image synthesis and decomposition framework that can synthesize rendering results based on input materials or predict the material channels of rendered images. IDArb~\cite{li2024idarb} combines cross-view and cross-component attention with illumination enhancement and view-adaptive training, enabling it to predict consistent material information from multiple input views.
These advances in material estimation methods significantly enhance the accuracy of inferring intrinsic material properties, offering an alternative to two-stage methods that first generate shaded images and then decompose them. In this work, we leverage the multi-view capabilities of IDArb~\cite{li2024idarb} to infer metallic and roughness channels, and design to model shaded image generation for integrating it with our modeling framework.

%% file: sec/3_method.tex
\section{Methods}
\label{sec:methods}
We aim to produce high-quality textured triangular meshes from untextured ones, guided by text or image prompts. Our pipeline is illustrated in Fig.~\ref{fig:framework}. In this section, we first outline the basic settings for multi-view generation as the preliminaries in Section~\ref{sec:preliminaries}. We then present our two key innovations: multi-channel multi-view generation in Section~\ref{sec:mcmv}, and agentic post-processing in Section~\ref{sec:agentic}.

\subsection{Preliminaries}
\label{sec:preliminaries}

Recent research in 3D texturing has shown increasing interest in leveraging multi-view generation with back-projection to accomplish the task~\cite{yang2024hunyuan3d,zhang2024clay,zhu2024mcmat}. 
The diffusion model learns the joint distribution of multi-view images $\bm x$, guided by intention inputs $\bm c$ (\ie, text or image prompts) and geometry control $\bm g$ (\eg, multi-view depth). The models follow the basic diffusion training method, where a timestep $t$ is sampled and predict the added noises $\bm \epsilon$ from noisy inputs $\bm x_t$ with model parameters $\bm \theta$. And the geometry information is injected with ControlNet~\cite{zhang2023adding}. The training process is illustrated as:
\begin{align}
\label{eq:cond_ldm_loss}
L(\bm \theta, \mathcal X_{MV}) = \mathbb E_{\bm x, \bm g, \mathbf{c}, t, \bm \epsilon}\Big[
\Vert \bm \epsilon - \bm{\epsilon_\theta}(\bm x_{t}; \bm g,\bm c,t) \Vert_{2}^{2}
\Big].
\end{align}

To handle PBR parameters, we can increase the number of channels in the multi-view images $\bm x$, inflating the RGB channels to support additional modalities. For instance, we can simply set the concatenated images, \ie, \{albedo, metallic, roughness\}, as the modeling target $\bm x$. Typically, we use latent diffusion models~\cite{rombach2022high}, where we can inflate the single-channel metallic/roughness to three channels and encode them before concatenation. Additionally, to process these multi-channel targets, we can either expand the outermost blocks~\cite{zhang2024clay} or create multiple branches from scratch~\cite{zhu2024mcmat}.
The architectural design can be based on multi-view consistency modules, such as pseudo-3D modules~\cite{shi2023mvdream} and additional 3D consistency adapters~\cite{huang2024mv}. Given the feature $\bm f$, a standard implementation involves acquiring the multi-view feature $\bm f'$ through a reshape operation and multi-view modules, as follows:
\begin{align}
\label{eq:multi-view}
\bm f' = {\rm Reshape} ({\rm Module}({\rm Reshape}(\bm f))).
\end{align}
After acquiring multi-view images, unlike earlier works~\cite{chen2023text2tex, richardson2023texture} to progressively produce new single-view images with RePaint~\cite{lugmayr2022repaint}, recent works choose to supply enough information in UV space and after UV unwarping, opt to inpaint with further UV inpainting or specially trained refinement networks~\cite{zhu2024mcmat, cheng2024mvpaint, zeng2024paint3d}. 

\subsection{Multi-Channel Multi-View Generation}
\label{sec:mcmv}

\noindent \textbf{Modeling Shaded and Albedo Images.}
We opt to model the multi-channel images of both multi-view shaded and albedo images. Existing research~\cite{zhang2024clay, zhu2024mcmat, huang2024material} attempts PBR modeling by jointly generating all PBR channels, \ie, albedo, metallic, and roughness. However, this strategy suffers from two key issues.
First, there is a significant domain gap between the PBR channels and the images without lighting control produced by the base foundation models, such as Stable Diffusion~\cite{rombach2022high}. This makes fine-tuning the multi-view models challenging. Even if we opt to train the multi-channel generation model from scratch, challenges persist due to the domain gap between the albedo and the remaining channels and the lack of strong generative priors, which are typically trained on large-scale 2D datasets like Laion~\cite{schuhmann2022laion}.
Second, the dataset of PBR materials is highly imbalanced. High-quality 3D asset datasets that include all three channels are scarce and valuable, whereas assets containing albedo maps are relatively more abundant and easier to curate with human labor in real industry settings.
In our method, rather than just producing the albedo multi-view images $\bm a$, we also include the shaded images $\bm s$ aimed at connecting to intrinsic decomposition models.

\noindent \textbf{Integration of Multi-View Intrinsic Decomposition.}
With intrinsic decomposition models specially trained on materials data, material decomposition can be performed from shaded images, indicating that material generation does not have to rely on a single-stage model. Recently, IDArb~\cite{li2024idarb} has achieved multi-view consistent decomposition with its use of attention mechanisms, which aligns well with our design.
In our approach, the generated shaded multi-view images serve as the input for IDArb to produce the material channels, \ie, metallic $\bm m$ and roughness $\bm r$:
\begin{align}
\label{eq:idarb}
\bm m, \bm r = {\rm IDArb}(\bm s).
\end{align}
This multi-stage process leverages the strengths of intrinsic decomposition models to generate high-quality PBR materials from more readily available shaded images, thus addressing domain gaps and data imbalances.

\noindent \textbf{Multi-View Design and Model Training.}
The effectiveness of fine-tuning multi-view generative models largely depends on the capabilities of the base model. As we aim to train additional layers to incorporate new channels, it is crucial to efficiently train the model while leveraging the foundational capabilities. To achieve this, we take a different approach from earlier works~\cite{zhang2024clay, cheng2024mvpaint}, which trained models on MVDream~\cite{shi2023mvdream} based on SD1.5 and reduced image resolution.
Instead, we design our training process using SDXL~\cite{podell2023sdxl}, a more advanced model, and integrate the MV-Adapter~\cite{huang2024mv}, which is trained on SDXL for multi-view consistency, through attention control:
\begin{equation}
\begin{aligned}
\label{eq:mvadapter}
    \bm{f}^{self} &= \text{SelfAttn}(\bm{f}^{in}) + \text{MultiViewAttn}(\bm{f}^{in}) + \\ &\text{ImageCrossAttn}(\bm{f}^{in}, \bm{f}^{ref}) + \bm{f}^{in},
\end{aligned}
\end{equation}
which incorporates self-attention and multi-view attention for querying input feature $\bm f^{in}$, as well as optional image cross-attention for the feature of image prompts $\bm f^{ref}$. 
The outermost input and output blocks are inflated to support the multi-channel design, following the standard operations also used in~\cite{zhang2024clay, liu2023hyperhuman}. We train these inflated blocks along with MV-Adapter for efficient training. 
We use rendered multi-view depth images for the geometry control and use ControlNet-Depth~\cite{zhang2023adding} trained on SDXL. To achieve better viewpoint coverage in the UV space, we use a training bundle of 6 views which includes 4 surrounding images with an azimuth gap of $90\degree$ and 2 images with elevation $\pm90\degree$.

\begin{figure*}[t]
    \centering
    \includegraphics[width=0.95\linewidth]{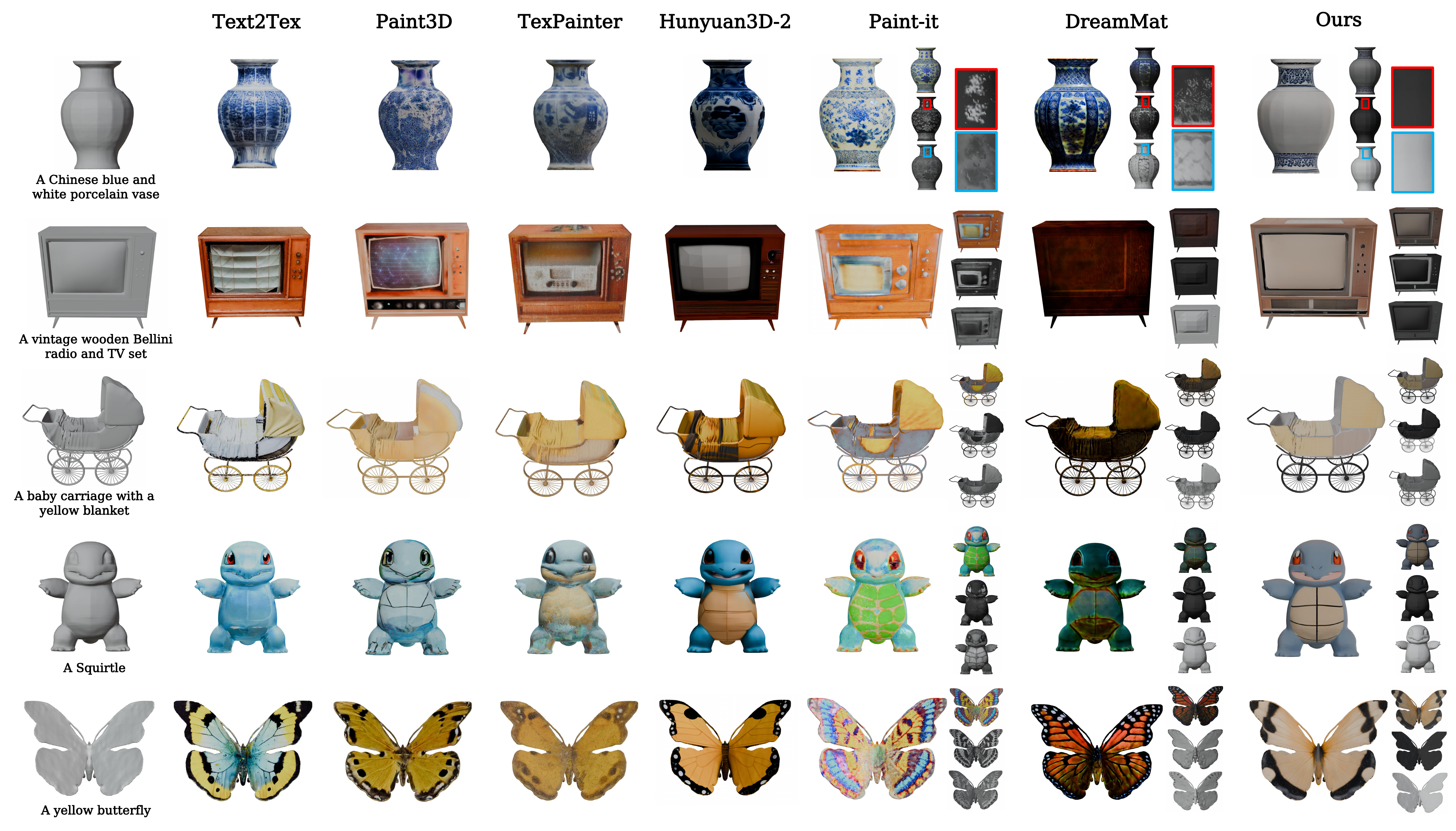}
    \caption{\textbf{Qualitative comparison on text-conditioned generation.} For methods that produce materials, we present the albedo, metallic, and roughness maps on the right of each case. The last two rows showcase the comparison results on generated meshes.
    }
    \label{fig:comp_text}
\end{figure*}

\begin{table*}[!t]
\begin{center}
\setlength{\tabcolsep}{4pt}
\caption{
\textbf{Quantitative comparison on text-conditioned generation.} The metrics with best performances are in bold.}
\vspace{-1em}
\resizebox{\textwidth}{!}{
\begin{tabular}{cccccccc}
\toprule 
\makecell{Metrics / Methods} & Text2Tex~\cite{chen2023text2tex} &
Paint3D~\cite{zeng2024paint3d} & TexPainter~\cite{zhang2024texpainter} & Hunyuan3D-2~\cite{zhao2025hunyuan3d} & Paint-it~\cite{youwang2024paint} & DreamMat~\cite{zhang2024dreammat} & \textbf{Ours} \\
\midrule
CLIP $\uparrow$  & 26.92   & 29.48   & 28.83   & 30.09  & 27.37  & 28.52  & \textbf{30.41} \\
FID $\downarrow$       & 128.76  & 110.51  & 121.85  & 106.91 & 117.04 & 114.03 &    \textbf{103.47} \\
KID $\downarrow$       & 21.44   & 14.58   & 17.76   & 13.07  & 16.01  & 18.45  &    \textbf{11.56} \\
Time $\downarrow$      & 10 min             & 3 min    & 8 min   & \textbf{75 sec}  & 40 min  & 1h 15 min   & 2 min \\
\bottomrule
\end{tabular}
}
\label{tab:quantitative_text}
\end{center}

\end{table*}

\subsection{Agentic Post-Processing}
\label{sec:agentic}

\noindent \textbf{Scoring Strategies.}
The intrinsic decomposition model also produces multi-view albedo images, which serve as a competing candidate. In our experiments, we observe that in approximately $13.7$\% of cases, the albedo produced by IDArb, $\bm a’$, is preferred over the generated one according to human preference. Therefore, we integrate GPT-4o~\cite{gpt4o} for scoring and select the one that achieves higher average ratings under five times of running to eliminate randomness, \ie, $\bm a = {\rm Select}({\rm Score}(\bm a), {\rm Score}(\bm a’))$. Since we produce multi-view images in each candidate set, we investigate three scoring strategies:

\begin{itemize}
    \item Scoring Once: We merge the generated albedo and the albedo decomposition of IDArb into one row each, and then concatenate the two lines of images into one image. We produce two scores according to this image in a single MLLM query.
    \item Scoring Each View: We score each view in each candidate set, and compute the average scores of each set.
    \item Scoring Each Set: We horizontally stitch the multi-view albedo images of 6 views into one image and query the MLLM twice for scores of each set. 
\end{itemize}
The score is out of 100. Based on our experiments, we perform Scoring Each Set strategy, due to its satisfactory performance and efficiency in token consumption.

\noindent \textbf{Best-of-N Selection.}
As the materials produced may not always align with the semantics of the objects controlled by text or image prompts, we optionally conduct a Best-of-N selection. Since the shaded images produced by the multi-view model serve as the input for IDArb, both modules may encounter instabilities in producing the correct materials. This approach performs better at the cost of increased computation time, where the multi-view model and IDArb can be run N times to prevent semantic errors.

%% file: sec/4_exp.tex
\section{Experiments}
\label{sec:exp}

\subsection{Implementation Details}
We utilize a rendered dataset processed from Objaverse~\cite{deitke2023objaverse}, selecting 6 views for each asset and rendering material channels, depth images, and shaded images under lighting conditions. This results in a comprehensive dataset comprising 220,000 rendered assets. For image-conditioned setting, we use the shaded images in the front view as the image prompts in training.
For the multi-view diffusion model, we modify the SDXL~\cite{podell2023sdxl} framework by integrating ControlNet-Depth~\cite{zhang2023adding} and MV-Adapter~\cite{huang2024mv}. We focus on training the inflated modules and MV-Adapter for 50,000 steps. To preserve image quality, we maintain the resolution at $768\times768$, as supported by the base model.
For scoring and selection, we employ GPT-4o~\cite{gpt4o}. In scenarios requiring multiple selections, we optionally set $N=3$. Detailed querying prompts are provided in the Appendix. 
After back-projecting the six views onto the UV plane, we use the image inpainting algorithm~\cite{telea2004image} to fill the gaps in the UV space, and then use Real-ESRGAN~\cite{wang2021real} to upsample the map by a factor of 2.

\subsection{Baselines and Evaluation Metrics}
We compare our method with several state-of-the-art methods for 3D appearance generation, including Text2Tex~\cite{chen2023text2tex}, TexPainter~\cite{zhang2024texpainter}, Paint3D~\cite{zeng2024paint3d}, Paint-it~\cite{youwang2024paint}, DreamMat~\cite{zhang2024dreammat}, and Hunyuan3D-2~\cite{zhao2025hunyuan3d}. Among these, DreamMat and Paint-it are capable of generating PBR material maps, while the other methods can only produce a single RGB texture map. Given that the mesh is provided, we utilize only the texture synthesis stage of Hunyuan3D-2 to generate the texture map with fixed geometry. Additionally, we compare our method with texturing methods that support image-conditioned generation, specifically Paint3D~\cite{zeng2024paint3d} and Hunyuan3D-2~\cite{zhao2025hunyuan3d}.
We take diverse 3D meshes from the Objaverse dataset~\cite{deitke2023objaverse} that
are not included in the training set for evaluation, as well as meshes generated by geometry generation model CraftsMan~\cite{li2024craftsman}, with 490 and 160 samples for each types of resources.
For the evaluation metrics, we calculated CLIP Score~\cite{radford2021learning}, FID~\cite{heusel2017gans}, and KID~\cite{binkowski2018demystifying} to quantitatively evaluate the generated appearance quality. The CLIP Score can reflect the matching degree between the generated texture and the input text, while the FID and KID are used to measure the difference between the generated images and the real images. We also include the time required to generate the texture maps to measure the efficiency of different methods.

\subsection{Experimental Results}

\begin{figure}[t]
\centering
\includegraphics[width=0.49\textwidth]{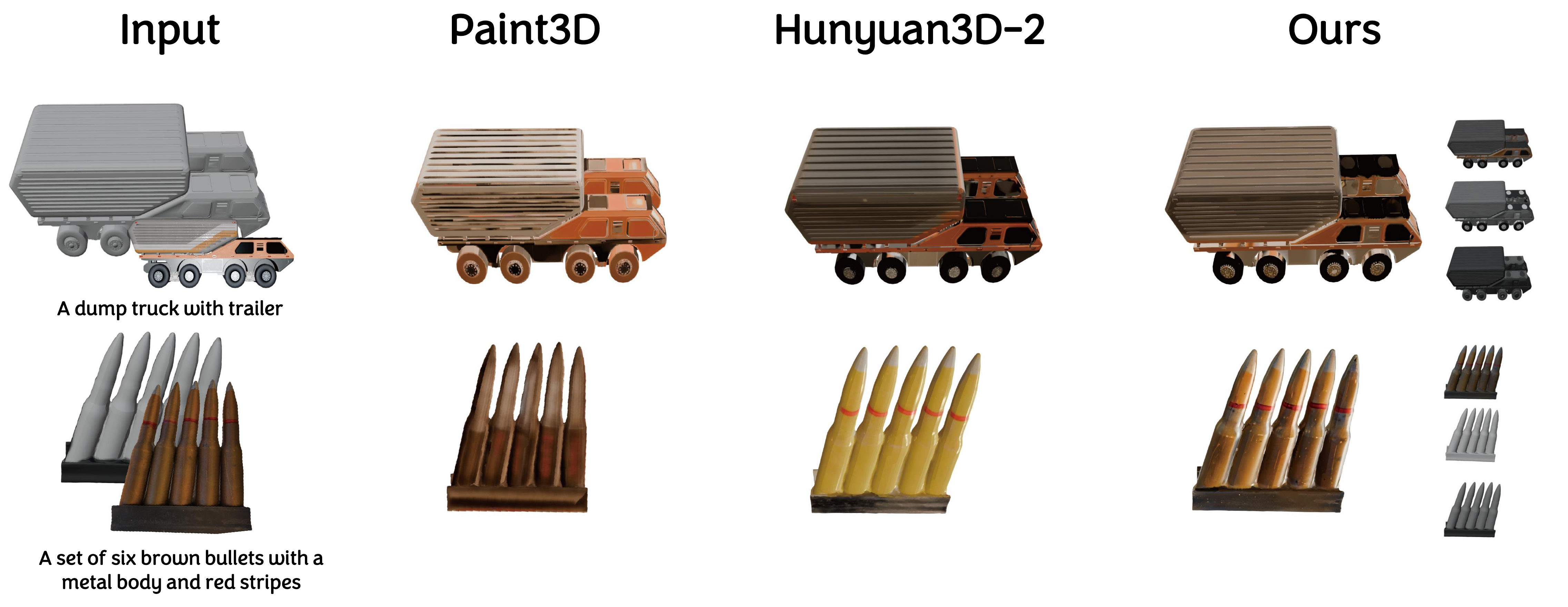}
\caption{\textbf{Qualitative comparison on image-conditioned generation.} Given the input mesh, along with the control text and image prompts, we sythesize PBR materials with high-fidelity.} 
\label{fig:comp_image}
\end{figure}

\noindent \paragraph{Qualitative Comparison.}
Fig.~\ref{fig:comp_text} and Fig.~\ref{fig:comp_image} present a qualitative comparison of 3D texturing in text-conditioned and image-conditioned settings. For methods that produce material outputs, we display the results along with the shaded images. The qualitative results demonstrate that our method achieves superior visual quality and PBR fidelity in the text-conditioned setting and closely matches Hunyuan3D-2 in the image-conditioned setting. Sometimes the PBR produced by baseline methods is messy, as illustrated in highlighted zones in Fig.~\ref{fig:comp_text}. The last two rows in Fig.~\ref{fig:comp_text} showcase the comparison results on generated meshes. It is observed that our results perform satisfactorily for both artist-created 3D models and generated models produced by CraftsMan~\cite{li2024craftsman}, indicating the generalizability of our method to imperfect meshes.

\noindent \paragraph{Quantitative Comparison.}
Table~\ref{tab:quantitative_text} demonstrates that our method achieves superior performance in text-conditioned texturing compared to baseline methods. This indicates that the results generated by our method are more consistent with the text prompts and visually closer to real images. Additionally, previous PBR generation methods~\cite{zhang2024dreammat, youwang2024paint} require at least tens of minutes for optimization, whereas our method only takes 2 minutes, significantly enhancing efficiency. For image-conditioned generation, Table~\ref{tab:quanlitative_image} shows that our method performs slightly poorer than Hunyuan3D-2, which excels in image-conditioned generation but does not produce material maps.

\begin{table}[t]
\begin{center}
\setlength{\tabcolsep}{4pt}
\caption{\textbf{Quantitative comparison on image-conditioned.}}
\vspace{-.8em}
\begin{tabular}{lccc}
\toprule 
Methods / Metrics & CLIP $\uparrow$ & FID $\downarrow$ & KID $\downarrow$ \\
\midrule
Paint3D~\cite{zeng2024paint3d} & 28.65 & 114.51 & 19.79 \\
Hunyuan3D-2~\cite{zhao2025hunyuan3d} & \textbf{30.38} & \textbf{96.62} & \textbf{10.51} \\
\textbf{Ours} & 29.23 & 108.27 & 16.09 \\
\bottomrule
\end{tabular}
\label{tab:quanlitative_image}
\end{center}
\end{table}

\begin{table*}[ht]
\begin{center}
\setlength{\tabcolsep}{3pt}
\caption{
\textbf{User study on text-conditioned generation.} The rating is of scale 1-5, where the higher the better.}
\vspace{-.8em}
\resizebox{\textwidth}{!}{
\begin{tabular}{lccccccc}
\toprule 
Metrics / Methods & Text2Tex~\cite{chen2023text2tex} &
Paint3D~\cite{zeng2024paint3d} & TexPainter~\cite{zhang2024texpainter} & Hunyuan3D-2~\cite{zhao2025hunyuan3d} & Paint-it~\cite{youwang2024paint} & DreamMat~\cite{zhang2024dreammat} & \textbf{Ours} \\
\midrule
Multi-View Consistency  & 2.64 & 2.70 & 3.11 & 3.45 & 3.18 & 3.02 & \textbf{3.67} \\
Textual Alignment       & 3.15 & 3.48 & 3.24 & 3.70 & 3.25 & 3.33 & \textbf{3.75} \\
Overall Quality         & 2.87 & 3.52 & 3.16 & 3.85 & 3.23 & 3.39 & \textbf{4.10} \\
\bottomrule
\end{tabular}
}
\label{tab:user_study}
\end{center}
\end{table*}

\subsection{User Study}
To further assess the visual quality of the generated texture maps, we conducted a comprehensive user study with 16 volunteers. For all baseline methods, we added the same ambient light to render the textured mesh from 24 fixed viewing angles. Each volunteer received 14 samples along with the corresponding text prompts. There were three evaluation indicators: overall quality, textual alignment, and multi-view consistency, each scored from 1 to 5, with 1 being the worst and 5 being the best. Table~\ref{tab:user_study} presents the results of our user study. In terms of multi-view consistency, textual alignment, and overall quality, our method received the highest scores. And we show the user study based on image input with Paint3D~\cite{zeng2024paint3d} and Hunyuan3D-2~\cite{zhao2025hunyuan3d} in Table~\ref{tab:user_study_img_cond}, different from the text-conditioned study, we use the reference alignment to measure the matching degree between the generated textures and the reference image. In addition, we conduct a user study comparing our method with DreamMat~\cite{zhang2024dreammat} and Paint-it~\cite{youwang2024paint} on the quality of PBR materials, as shown in Table~\ref{tab:user_study_pbr}, our method received the highest scores for albedo, metallic, and roughness quality.

\begin{table}[!t]
\begin{center}
\setlength{\tabcolsep}{3pt}
\caption{\textbf{User study on image-conditioned generation.}}
\vspace{-0.8em}
\resizebox{0.4\textwidth}{!}{
\begin{tabular}{lccc}
\toprule 
Methods / Metrics  & \makecell{Multi-View \\ Consistency} & \makecell{Reference \\ Alignment} & \makecell{Overall \\ Quality} \\
\midrule
Paint3D~\cite{zeng2024paint3d} & 2.94 & 2.76 & 3.23 \\
Hunyuan3D-2~\cite{zhao2025hunyuan3d} & 3.68 & \textbf{3.85} & \textbf{3.95} \\
\textbf{Ours} & \textbf{3.72} & 3.50 & 3.67 \\
\bottomrule
\end{tabular}
}
\label{tab:user_study_img_cond}
\end{center}
\end{table}

\begin{table}[!t]
\begin{center}
\setlength{\tabcolsep}{4pt}
\caption{\textbf{User study on text-conditioned PBR quality.}}
\vspace{-.8em}
\resizebox{0.47\textwidth}{!}{
\begin{tabular}{lccc}
\toprule 
Methods / Metrics & Albedo Quality & Metallic Quality & Roughness Quality \\
\midrule
DreamMat~\cite{zhang2024dreammat} & 2.93 & 3.45 & 3.60 \\
Paint-it~\cite{youwang2024paint} & 2.77 & 3.10 & 3.12 \\
\textbf{Ours} & \textbf{4.05} & \textbf{3.84} & \textbf{3.75} \\
\bottomrule
\end{tabular}
}
\label{tab:user_study_pbr}
\end{center}
\end{table}

\subsection{Ablation Study}
In this section, we include the ablation study in various designs, presented in Table.~\ref{tab:ablation}. We conduct all the experiments on text-conditioned generation.

\begin{table}[h]
\begin{center}
\setlength{\tabcolsep}{4pt}
\caption{\textbf{Ablation study.}}
\vspace{-.8em}
\resizebox{0.47\textwidth}{!}{
\begin{tabular}{lccc}
\toprule 
Methods / Metrics & CLIP $\uparrow$ & FID $\downarrow$ & KID $\downarrow$ \\
\midrule
Joint Modeling & 28.56 & 114.20 & 16.07 \\
Hunyuan+IDArb & 27.25 & 119.64 & 17.81 \\
w/o Scoring and Selection & 29.93 & 105.76 & 12.38 \\
\textbf{Full Model (Ours)} & \textbf{30.41} & \textbf{103.47} & \textbf{11.56} \\
\textbf{Full Model + Best-of-N (Ours)}&  
\textbf{30.57} & \textbf{102.61} & \textbf{11.34}  \\

\bottomrule
\end{tabular}
}
\label{tab:ablation}
\end{center}
\end{table}

\begin{figure}[h]
\centering
\includegraphics[width=0.49\textwidth]{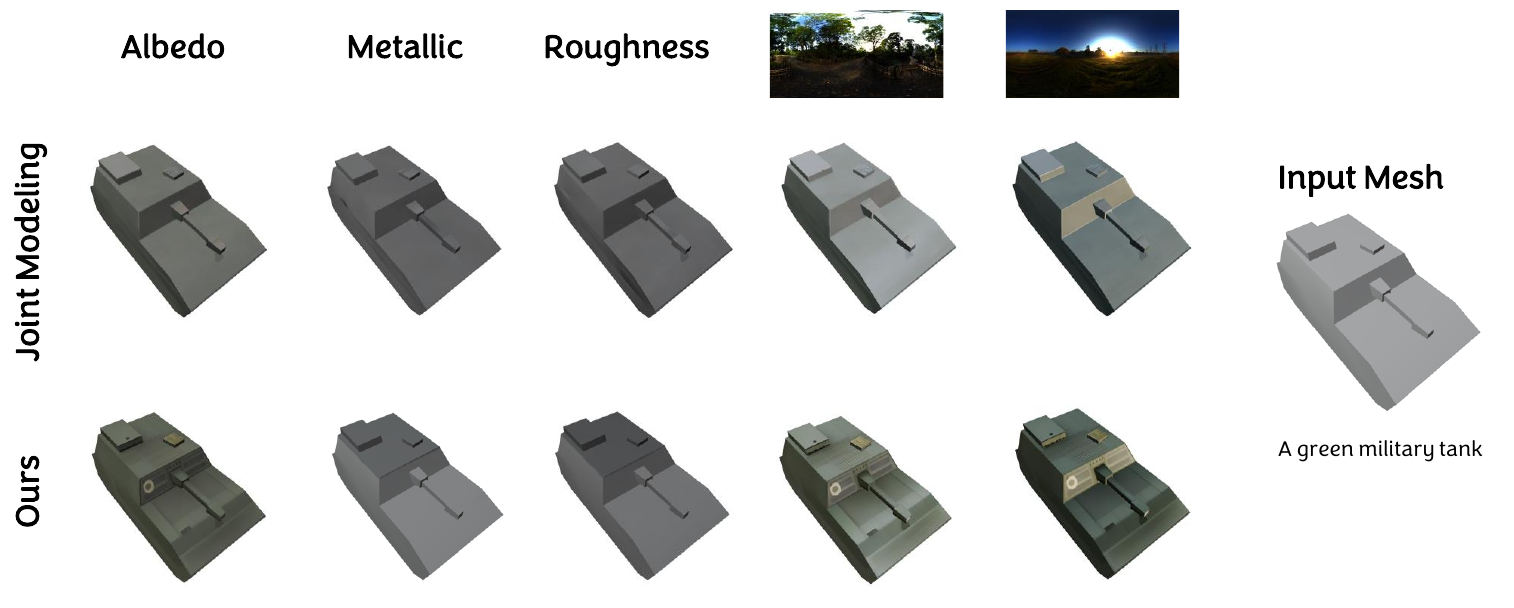}
\caption{\textbf{Comparison on text-conditioned PBR texturing of Joint Modeling and Ours.} We compare the PBR materials and the rendering results under two environment light maps.} 
\label{fig:comparison_joint_modeling}
\end{figure}

\begin{figure}[h]
\centering
\includegraphics[width=0.49\textwidth]{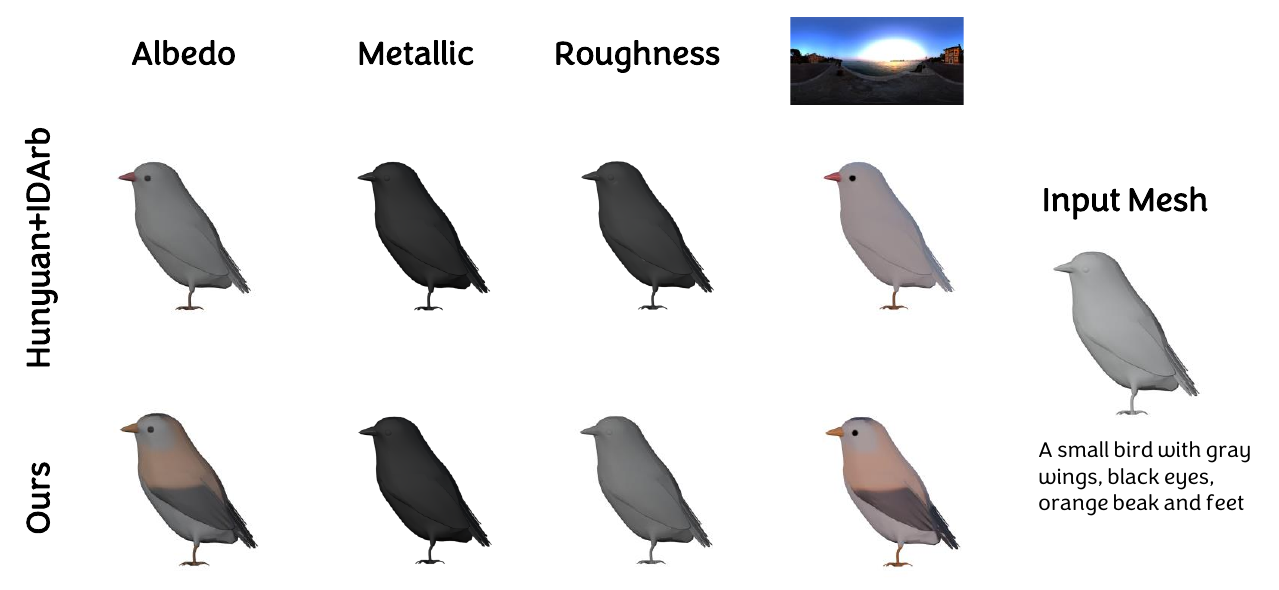}
\caption{\textbf{Comparison on text-conditioned PBR texturing of Hunyuan+IDArb and Ours.}} 
\label{fig:comparison_hunyuan_plus_idarb}
\end{figure}

\begin{figure}[h]
\centering
\includegraphics[width=0.49\textwidth]{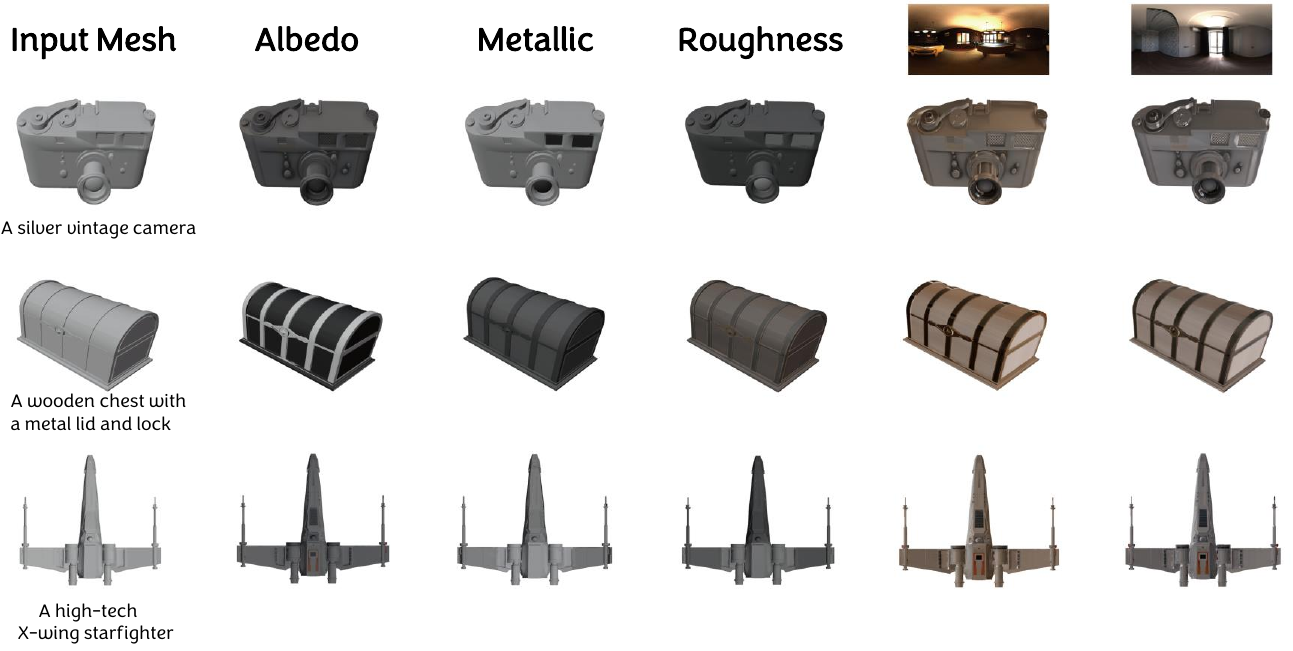}
\caption{\textbf{Results of our generated PBR textures and relighting performances.} We select two types of environment lighting maps.} 
\label{fig:relighting}
\end{figure}

\begin{figure*}[t]
    \centering
    \includegraphics[width=0.99\linewidth]{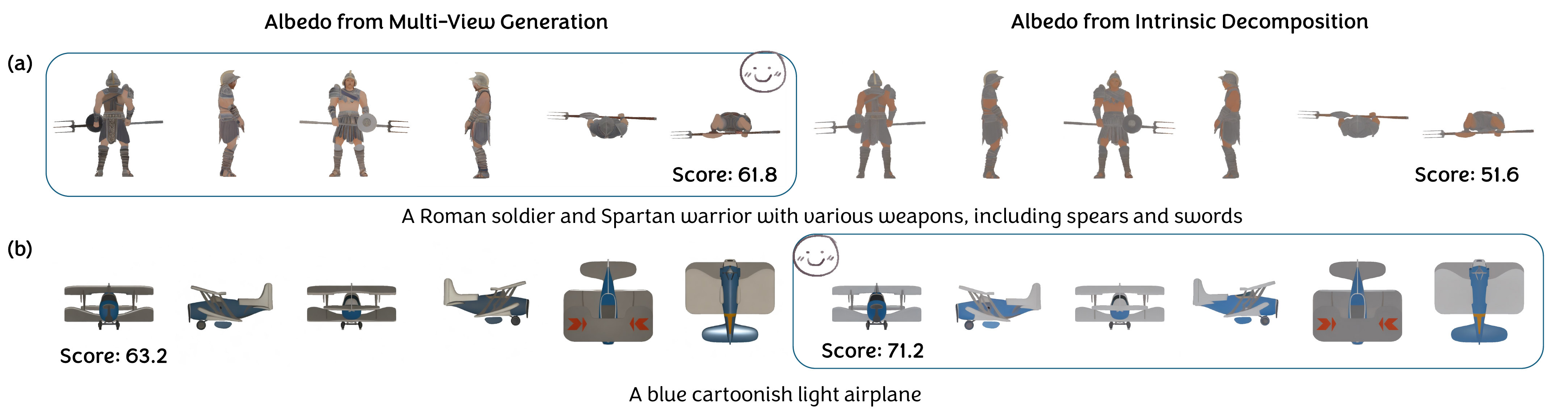}
    \caption{\textbf{Illustration of MLLM scoring and selection.} MLLM helps find the better albedo from multi-view generation and intrinsic decomposition. \textbf{(a)} Intrinsic decomposition produces blurry albedo. \textbf{(b)} Multi-view generation produces albedo with lighting effects.
    }
    \label{fig:mllm_selection}
\end{figure*}

\noindent \paragraph{Effectiveness of Dual Channel Modeling.}
One of our key design choices is to avoid generating all PBR materials in a joint modeling manner, as this approach suffers from difficult training and imbalanced data. To illustrate this point fairly, we train a model with a similar architecture to ours, incorporating SDXL and MV-Adapter, but using additional inflated blocks for the extra modalities. The training was based on rendering datasets covering assets that contain all three channels, totaling about 80,000 assets, which is fewer than the dataset comprising only albedo images. The quantitative and qualitative results, termed as Joint Modeling, are presented in Table~\ref{tab:ablation} and Fig.~\ref{fig:comparison_joint_modeling}, supporting our argument.
Additionally, our method produces significantly better albedo compared to the albedo generated by Joint Modeling, which may produce unsatisfactory details. This is illustrated in Fig.~\ref{fig:comparison_joint_modeling}.

\noindent \paragraph{Effectiveness of Intrinsic Decomposition Design.}
In MuMA, we focus on modeling shaded and albedo images, where the shaded images can be connected to intrinsic decomposition models for material channels. Apart from Joint Modeling, another straightforward method is to acquire materials via the generated images and material decomposition models. Therefore, we use the state-of-the-art method, Hunyuan3D-2, along with IDArb, to acquire materials and render the shaded images for comparison. As shown in Table~\ref{tab:ablation}, our method outperforms both Joint Modeling and Hunyuan+IDArb by a large margin.
It is noted that Hunyuan+IDArb produces poorer results than Hunyuan3D-2 because IDArb generates poor albedo from the images produced by Hunyuan3D-2. 
We include the comparison in Fig.~\ref{fig:comparison_hunyuan_plus_idarb}. The generated images of Huyuan3D-2 is used to serve as the input of IDArb and due to the domain gap, the albedo exhibits poor visual quality and does not align with the text prompt.
Additionally, it is infeasible to directly treat the generated images of Hunyuan3D-2 as the albedo since they contain lighting cues. Combining the dual channel modeling and intrinsic decomposition design, our method enables high-quality PBR texturing and facilitates relighting in various lighting environments, shown in Fig.~\ref{fig:relighting}.

\noindent \paragraph{Effectiveness of Agentic Design and Comparison of Scoring Strategies.}
We include an ablation study on scoring and selection with MLLM in Table~\ref{tab:ablation}. It is observed that this method clearly improves performance at a relatively low cost, as we have two candidate sets of albedo images. 
The illustration of the effectiveness of selection is shown in Fig.~\ref{fig:mllm_selection}, where sometimes the albedo produced by IDArb can be competitive.
We compare three scoring strategies in Table~\ref{tab:strategy_comparison}. While Scoring Once strategy requires only one MLLM query and consumes the fewest tokens, we select the Scoring Each Set strategy due to its higher performance and limited additional token usage. We also present the results with the optional Best-of-N strategy in Table~\ref{tab:ablation}, which performs better at the expense of increased computation time.

\begin{table}[t]
\begin{center}
\setlength{\tabcolsep}{4pt}
\caption{\textbf{Comparison on different scoring strategies of MLLM.}}
\vspace{-.8em}
\resizebox{0.47\textwidth}{!}{
\begin{tabular}{lcccc}
\toprule 
Methods / Metrics & CLIP $\uparrow$ & FID $\downarrow$ & KID $\downarrow$ & Tokens $\downarrow$ \\
\midrule
Scoring Once     & 30.24  & 104.85 & 11.91 & \textbf{1.00x} \\
Scoring Each View & 30.11  & 105.23 & 12.02 & 1.80x \\
\textbf{Scoring Each Set}  & \textbf{30.41} & \textbf{103.47} & \textbf{11.56} & 1.16x \\
\bottomrule
\end{tabular}
}
\label{tab:strategy_comparison}
\end{center}
\end{table}

%% file: sec/X_suppl.tex
\clearpage
\maketitlesupplementary

\subsection{Scoring Prompt Templates}
To fulfill the scoring task in the agentic system, we use GPT-4o~\cite{gpt4o} to produce ratings based on generated images. We include the prompts of different scoring strategies and Best-of-N strategy in this section. Here \textbf{I} indicates the input generated image, \textbf{I$_{m/r}$} indicates the metallic or roughness image, and \textbf{T} is the text description.

\begin{table}[h!]
\centering
\begin{minipage}{1.0\columnwidth}
\vspace{0mm}    
\centering
\begin{tcolorbox}
\footnotesize

\textbf{(a) Scoring Once} \\

Please rate the following image based on the prompt: \\
Image (Base64 encoded): \textit{Multi-view albedo image [\textbf{I}]} \\
Prompt: \textit{Multi-view generated albedo images of the [\textbf{T}]} \\
The first line and the second line correspond to two different methods respectively. Please provide a score out of 100 for each method, the higher the better. \\

\textbf{(b) Scoring Each View} \\

Please rate the following image based on the prompt: \\
Image (Base64 encoded): \textit{Albedo image [\textbf{I}]} \\
Prompt: \textit{Generated albedo image of the [\textbf{T}]} \\
Please provide a score out of 100, the higher the better. \\

\textbf{(c) Scoring Each Set} \\

Please rate the following image based on the prompt: \\
Image (Base64 encoded): \textit{Multi-view albedo image [\textbf{I}]} \\
Prompt: \textit{Multi-view generated albedo images of the [\textbf{T}]} \\
Please provide a score out of 100, the higher the better.
\end{tcolorbox}

\vspace{-2mm}
\caption{\textbf{The prompt of querying GPT-4o for three types of scoring strategies.}}
\label{prompt}
\end{minipage}
\end{table}

\begin{table}[h!]
\centering
\begin{minipage}{1.0\columnwidth}
\vspace{0mm}    
\centering
\begin{tcolorbox}
\footnotesize

\textbf{Scoring Metallic and Roughness} \\

Please rate the following image based on the prompt: \\
Image (Base64 encoded): \textit{Multi-view metallic/roughness image [\textbf{I$_{m/r}$}]} \\
Prompt: \textit{Multi-view generated metallic/roughness images of the [\textbf{T}]} \\
Please provide a score out of 100, the higher the better.
\end{tcolorbox}

\vspace{-2mm}
\caption{\textbf{The prompt of querying GPT-4o for choosing metallic and roughness in Best-of-N selection.}}
\label{rm_prompt}
\end{minipage}
\end{table}

\subsection{Limitations}
\noindent\textbf{Multi-view Inconsistency.} For most 3D models with simple geometric structures, our generative model can produce six views that are consistent from multiple perspectives. However, for more complex input meshes, the generated views may exhibit inconsistencies, particularly in the top and bottom perspectives, which are prone to incorrect colors. This issue arises due to the significant view distribution gap between elevation = $0\degree$ and elevation = $\pm90\degree$, causing the cross-view attention mechanism between the top/bottom and horizontal perspectives to be less effective. In the future, we plan to include more training data, which may help alleviate this problem.

\noindent\textbf{Inaccurate Material.} When encountering samples not present in the training data, our generative model sometimes produces albedo and rendering images with incorrect lighting effects. For instance, certain areas of the albedo image may still exhibit lighting effects, or some parts of the rendering image may display abnormal lighting conditions. Feeding these abnormal renderings into IDArb~\cite{li2024idarb} can result in incorrect metallic and roughness values. Additionally, IDArb itself can experience errors in material decomposition. To address this issue, we may need to incorporate ambient light information during the training of our generative model to enhance its perception of lighting. Furthermore, considering the inputs of material decomposition models like IDArb~\cite{li2024idarb, zeng2024rgb, hong2024supermat} are all single-modal images, adding additional prompt texts for multi-modal control may help improve the accuracy of material prediction.

\noindent\textbf{Texture Inpainting in Unobserved Areas.} Since the generated six views cannot completely cover the entire 3D model and self-occlusion often occurs in complex meshes, there will be gaps in the UV map obtained after back-projection. We currently fill these areas on the UV plane using the color of neighboring points, which may lead to inconsistent textures in 3D space. To address this issue, training a diffusion model specifically for UV inpainting~\cite{bensadoun2024meta, zeng2024paint3d, zhu2024mcmat} may be a better solution.

\begin{figure*}[h]
    \centering
    \begin{tabular}{cc}
    \includegraphics[width=0.48\linewidth]{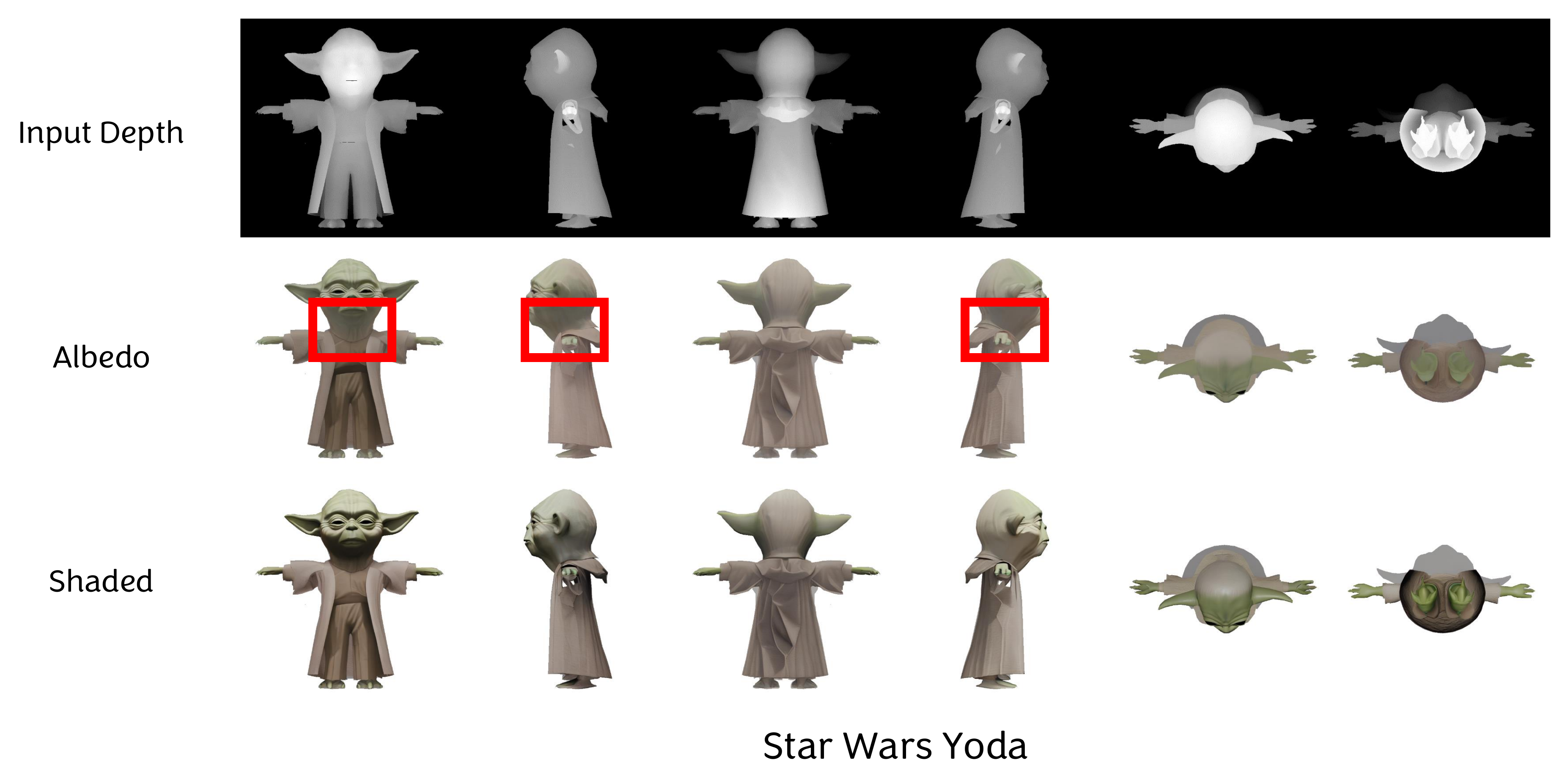}
    & \includegraphics[width=0.48\linewidth]{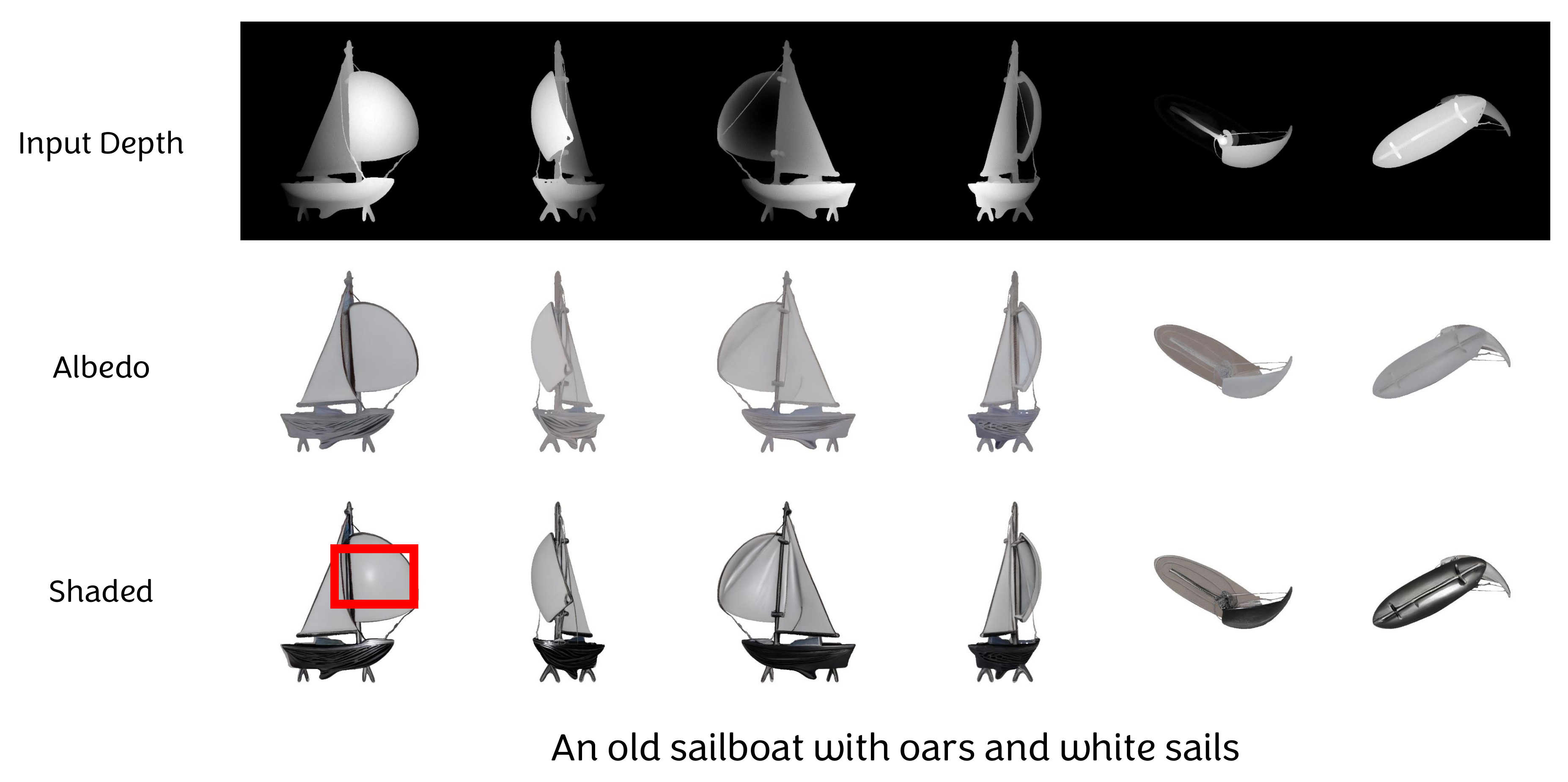}\\
    \includegraphics[width=0.48\linewidth]{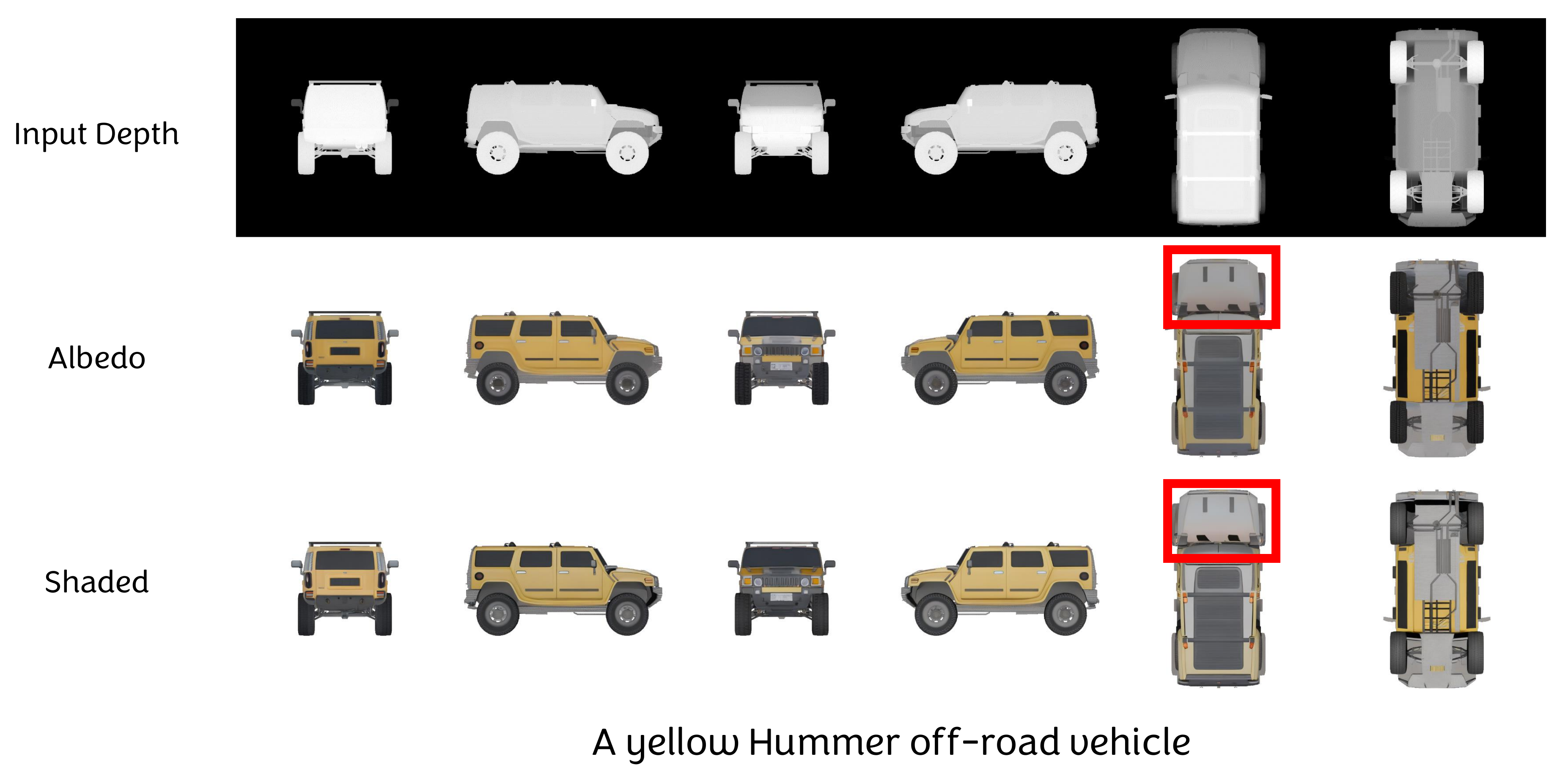} 
    & \includegraphics[width=0.48\linewidth]{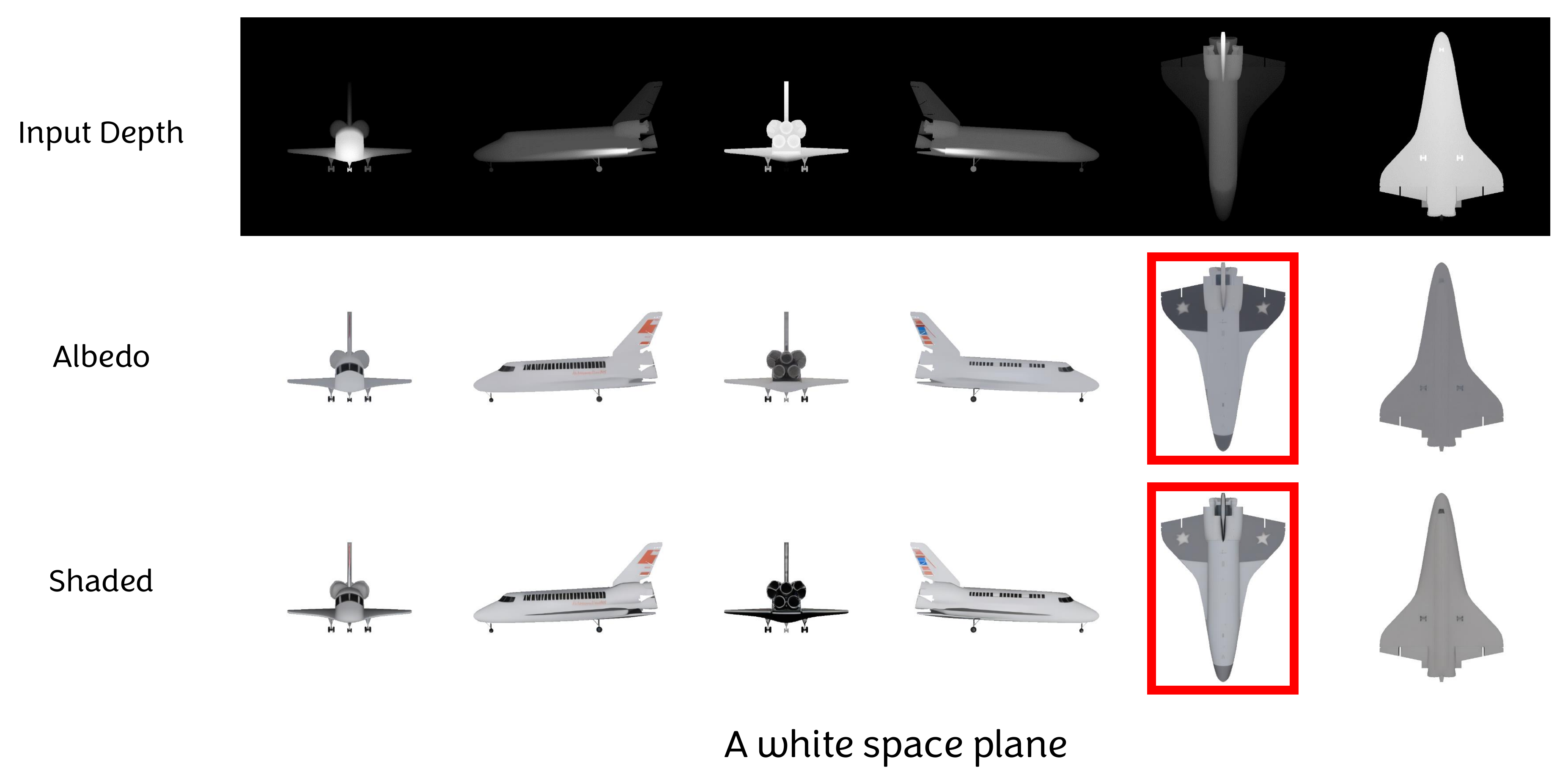} \\
    \end{tabular}
    \vspace{-.5em}
    \caption{\textbf{Failure cases on multi-view generation.} The multi-view generation mainly produces three types of imperfect results, \ie, shadow effects, incorrect semantics on lighting effects, and inconsistent colors.}
    \label{fig:failure_multiview}
\end{figure*}

\subsection{Failure Cases}
We present failure cases of multi-view generation in Fig.~\ref{fig:failure_multiview}. These failures can be categorized into three main types. First, the albedo images may exhibit shadow effects, likely due to imperfect training data. This issue is highlighted in the red rectangle in the first case, ``Star Wars Yoda''. Second, the shaded images may display incorrect lighting reflections. For instance, in the sailboat case, the sail shows an excessively strong reflection, which is inconsistent with the material properties. Third, the color may be imperfect in views with elevation angles of $\pm 90\degree$. As shown in the last two cases, the vehicle and the space plane, the colors within the red rectangles do not align perfectly with the surrounding views.

\subsection{More Results}
\begin{figure*}[th]
    \centering
    \begin{tabular}{c}
    \includegraphics[width=0.85\linewidth]{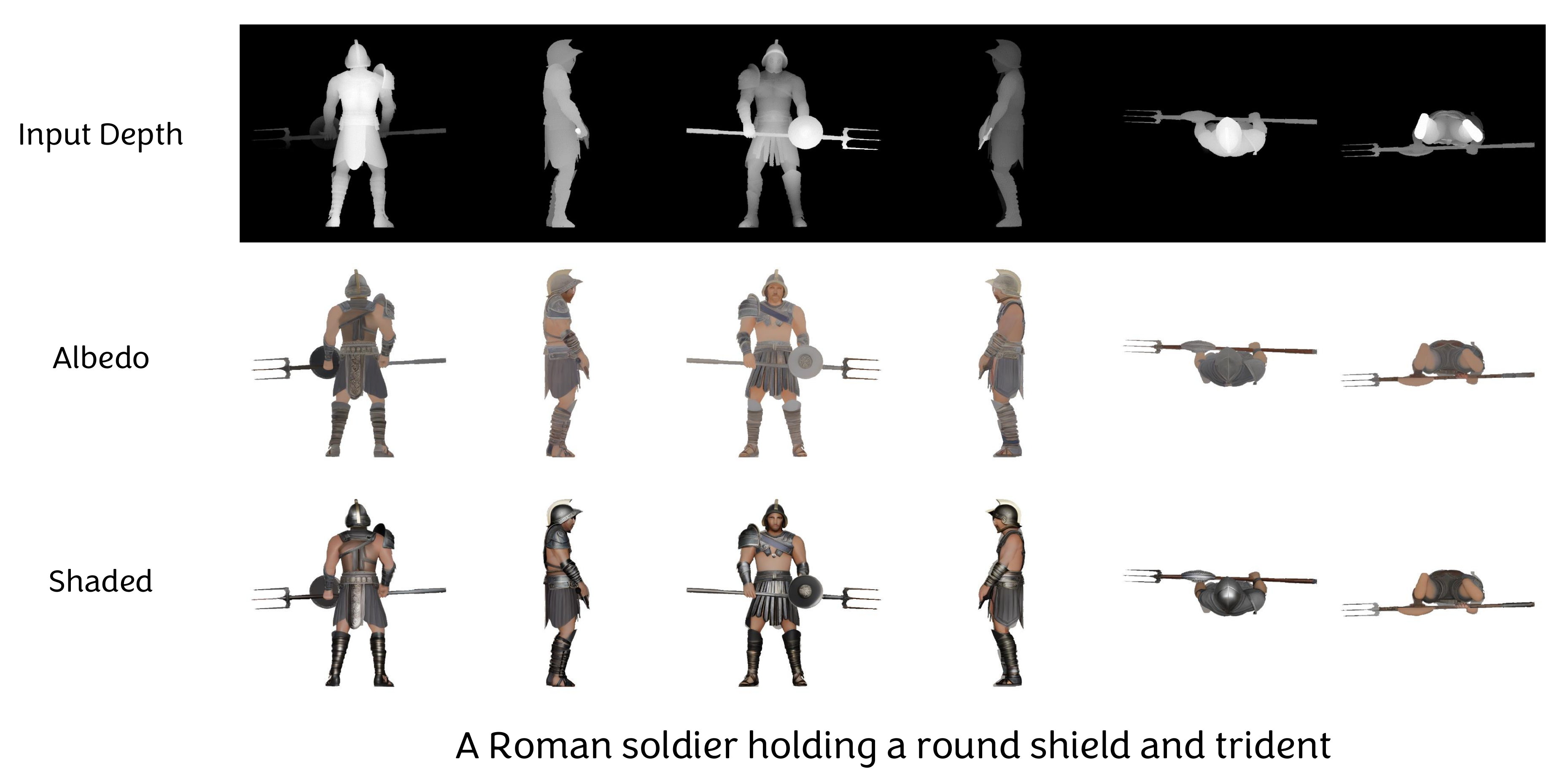} \\
    \includegraphics[width=0.85\linewidth]{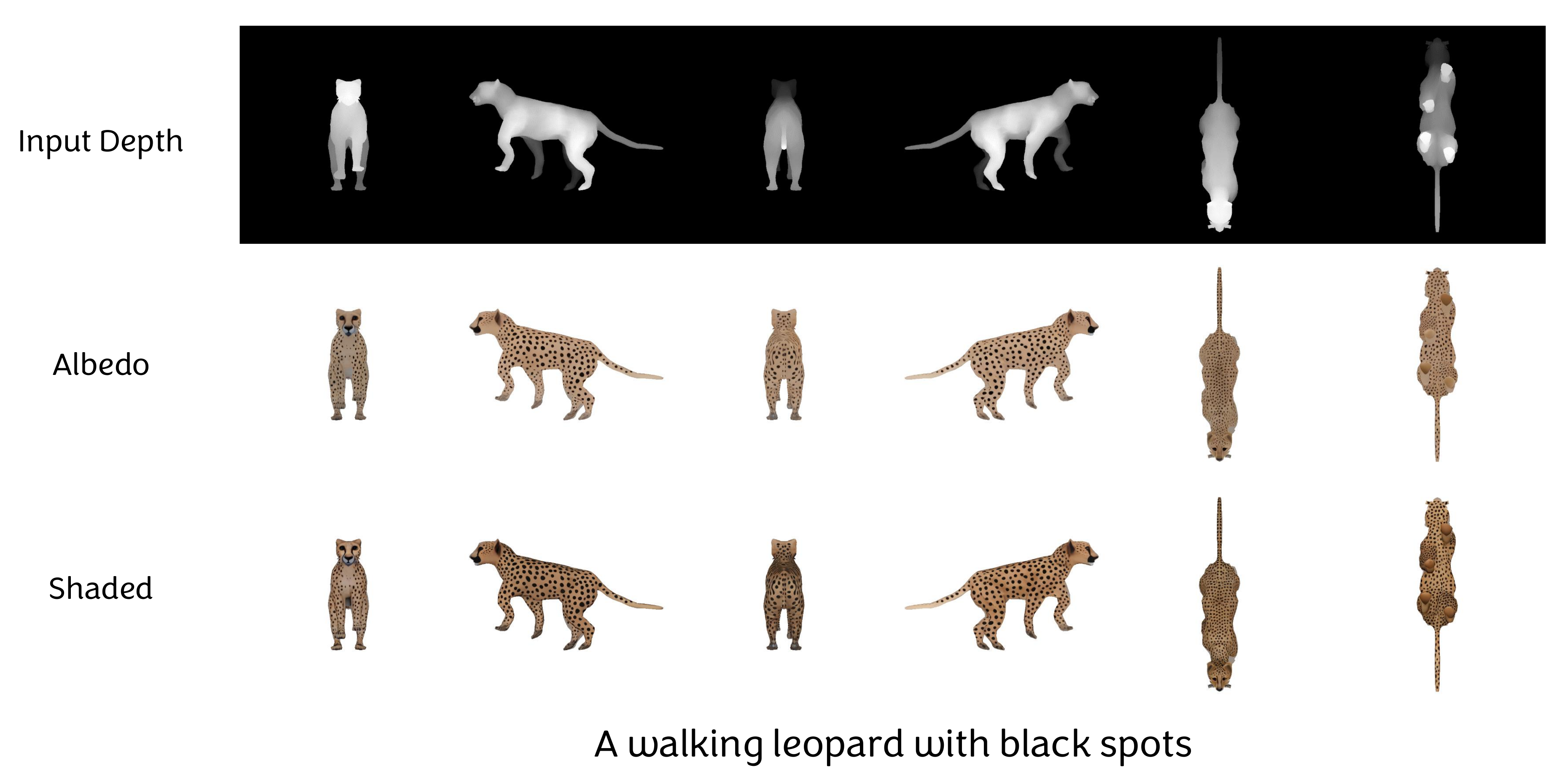} \\
    \includegraphics[width=0.85\linewidth]{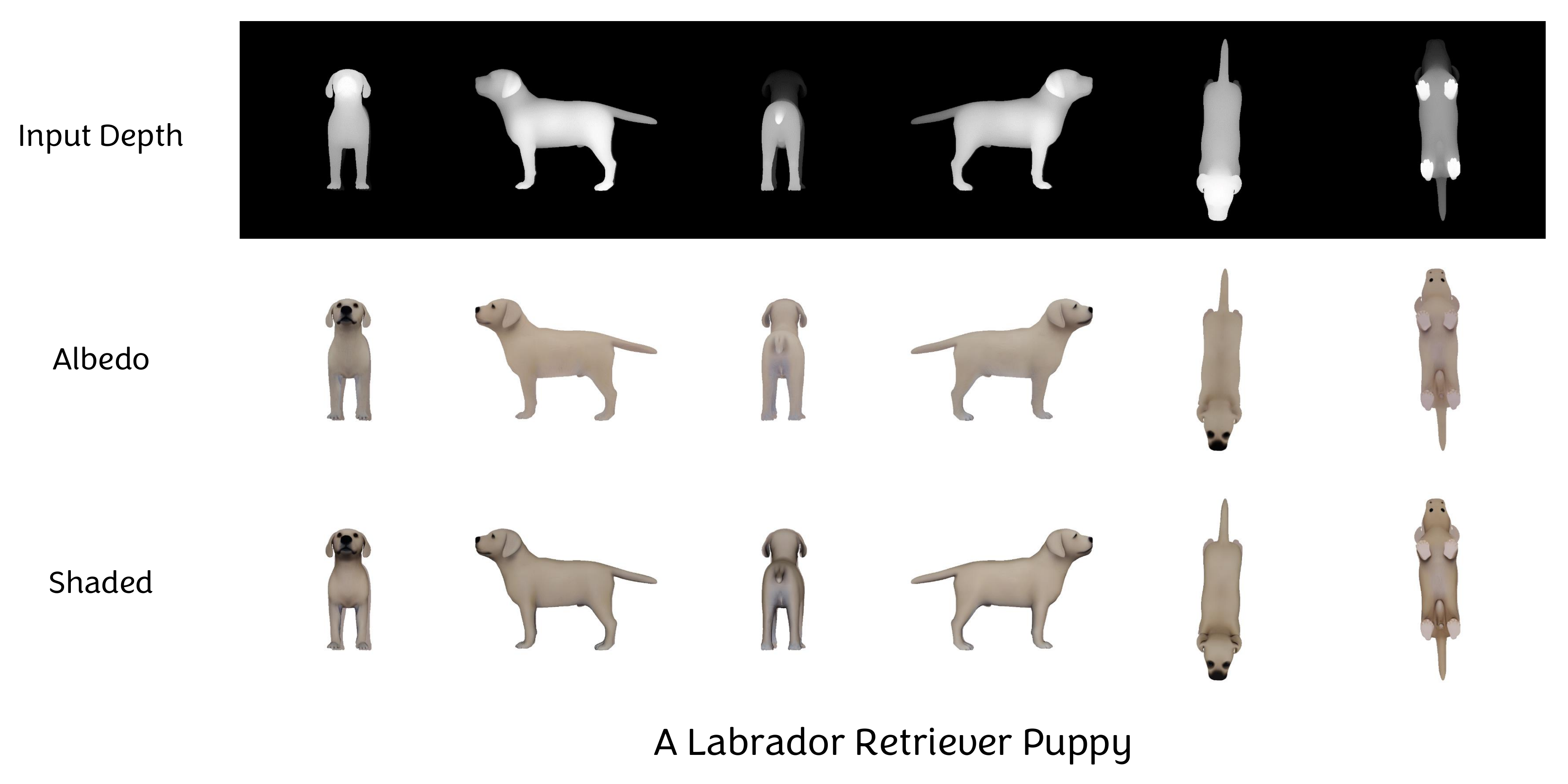} \\
    \end{tabular}
    \vspace{-0.5em}
    \caption{\textbf{Generated multi-view images of MuMA.}}
    \label{fig:more_multiview1}
\end{figure*}

\begin{figure*}[th]
    \centering
    \begin{tabular}{c}
    \includegraphics[width=0.85\linewidth]{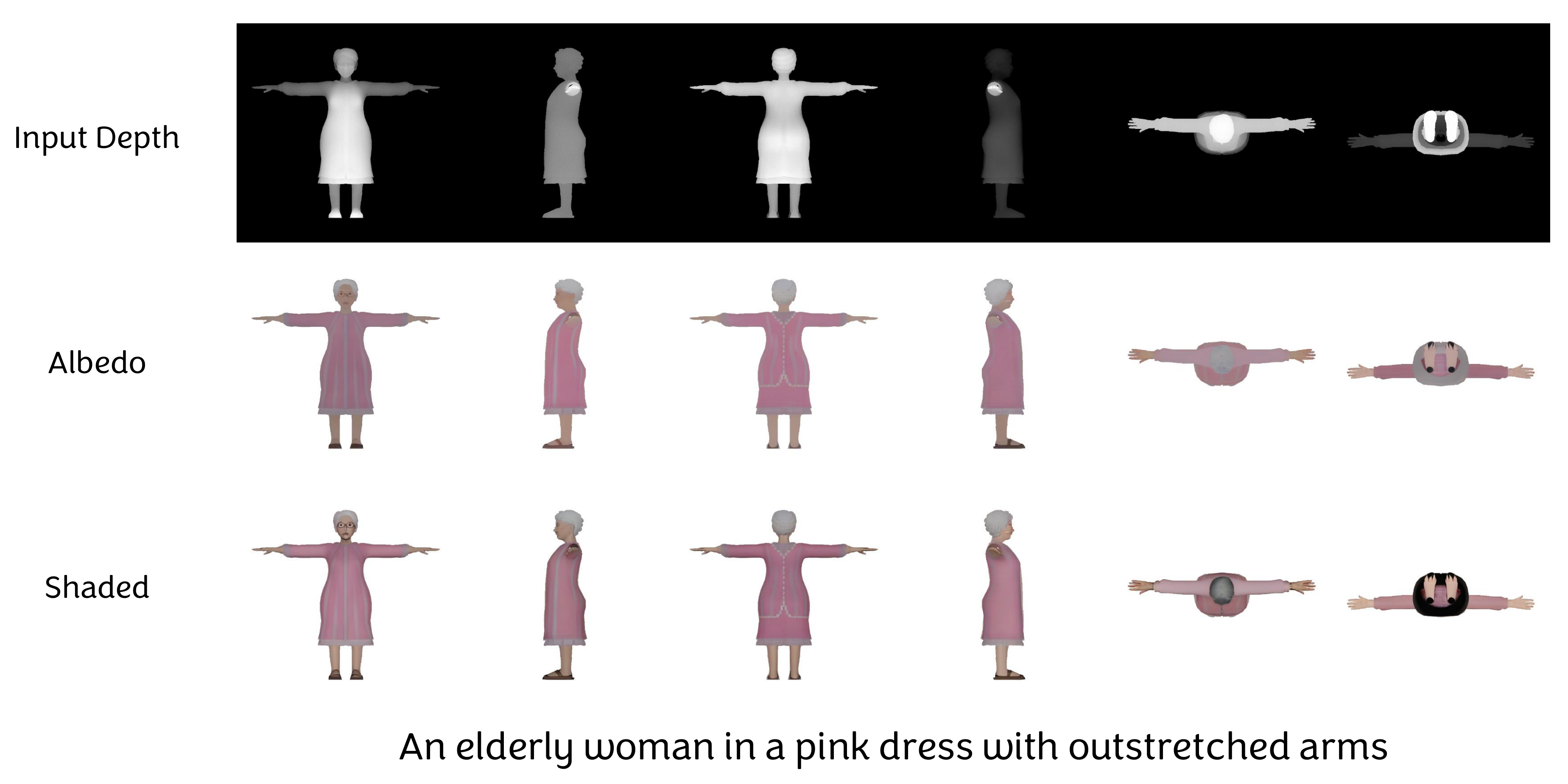} \\
    \includegraphics[width=0.85\linewidth]{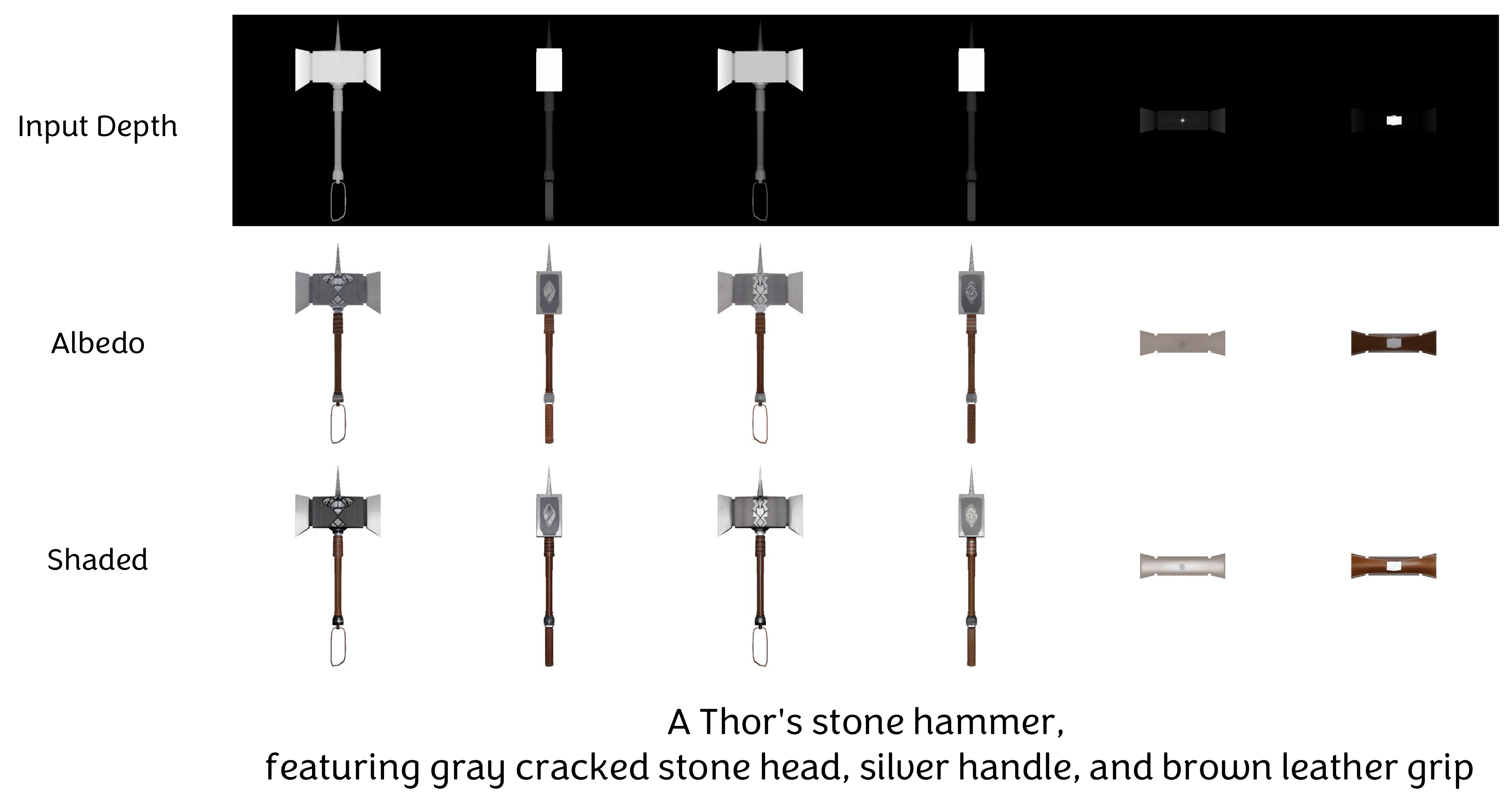} \\
    \end{tabular}
    \vspace{-0.5em}
    \caption{\textbf{Generated multi-view images of MuMA (Cont'd).}}
    \label{fig:more_multiview2}
\end{figure*}

\begin{figure*}[h]
\centering
\includegraphics[width=0.99\linewidth]{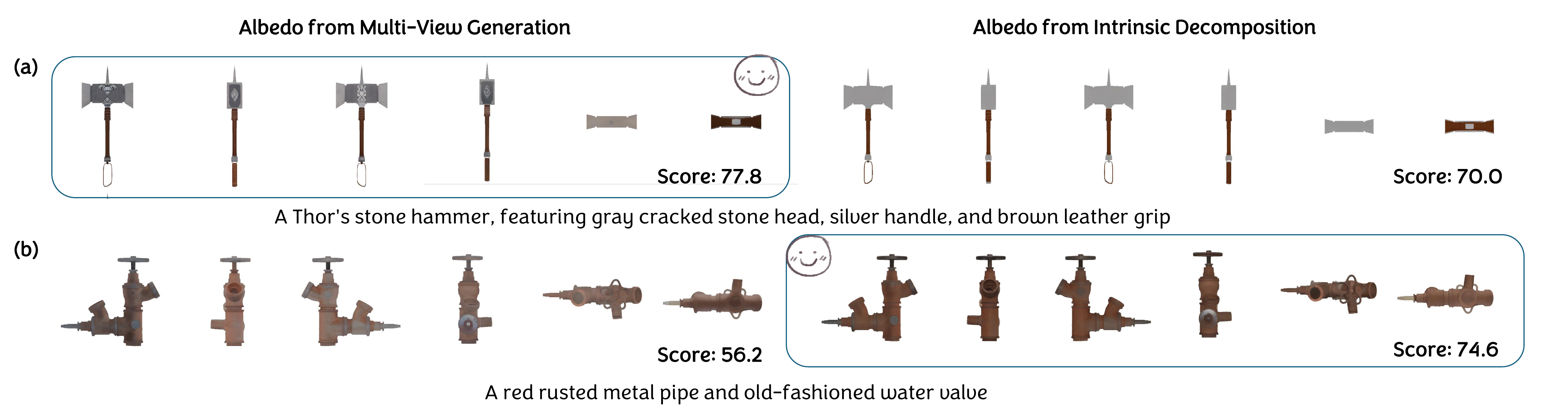}
\caption{\textbf{More results on MLLM scoring and selection.} \textbf{(a)} Intrinsic decomposition fails to produce great details. \textbf{(b)} Multi-view generation produces multi-view inconsistency in color.} 
\label{fig:more_scoring_selection}
\end{figure*}

\begin{figure*}[h]
\centering
\includegraphics[width=0.99\linewidth]{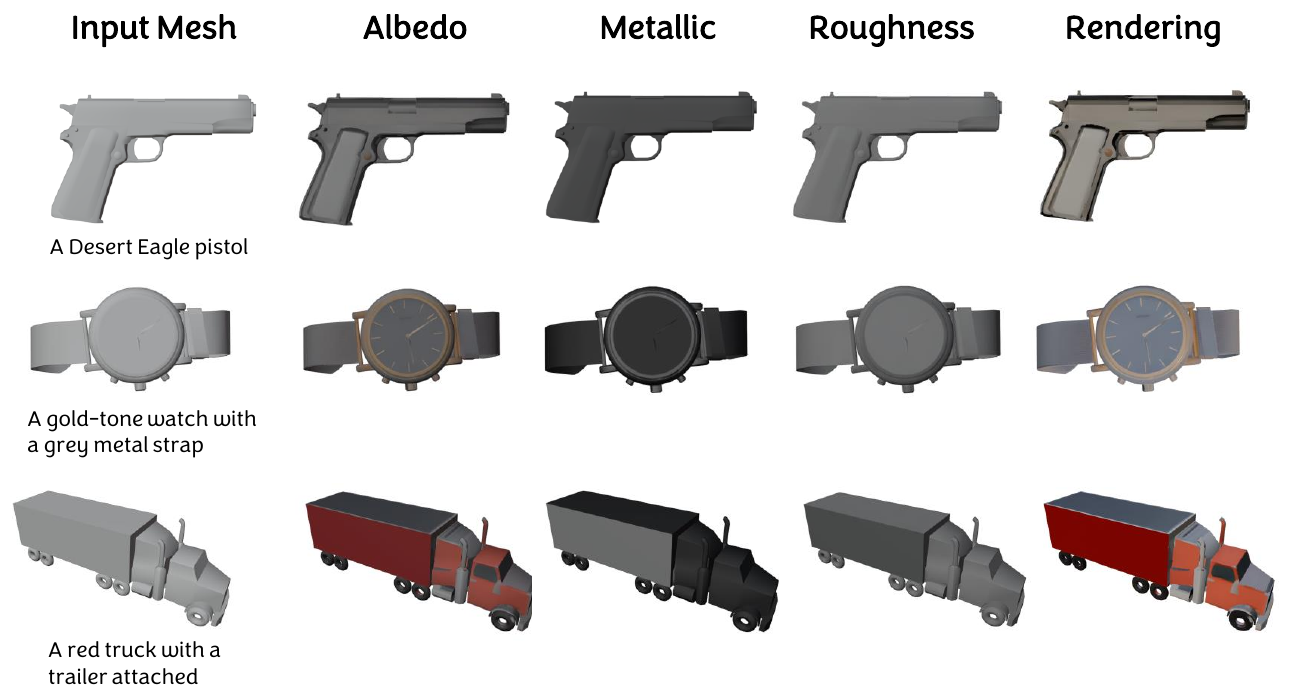}
\caption{\textbf{More results on 3D texturing.} We select a fixed environment lighting map for rendering the shaded images.} 
\label{fig:more_3d_texturing}
\end{figure*}

We present multi-view generation results in Fig.~\ref{fig:more_multiview1} and Fig.~\ref{fig:more_multiview2}. The results on text-conditioned multi-view generation is satisfactory, which is core in the system, given the rendered depth maps of the untextured meshes. We include two more cases on MLLM scoring and selection in Fig.~\ref{fig:more_scoring_selection}. It is observed that MLLM scoring and selection helps find the best between multi-view generation and intrinsic decomposition, leading to better details and consistency. Finally, we present three more cases on 3D texturing with rendering, shown in Fig.~\ref{fig:more_3d_texturing}.

